\documentclass{article} %
\usepackage[final]{colm2026_conference}

\usepackage{microtype}
\usepackage{hyperref}
\usepackage{url}
\usepackage{booktabs}

\usepackage{graphicx}
\usepackage[T1]{fontenc}   %
\usepackage{amsmath}
\usepackage{listings}      %
\usepackage{longtable}     %
\usepackage{xcolor}

\definecolor{darkblue}{rgb}{0, 0, 0.5}
\hypersetup{colorlinks=true, citecolor=darkblue, linkcolor=darkblue, urlcolor=darkblue}

\title{Reward-Driven Collider Physics Challenges for Self-Improving Agentic AI via Reinforcement Learning}

\author{Alexander Yue \\ 
Stanford University \\
450 Jane Stanford Way, Stanford, CA 94305 \\
\texttt{alexyue@stanford.edu} \\
\And
Qibin Liu \\ 
SLAC National Accelerator Laboratory \\
Menlo Park, CA 94025 \\
\texttt{qibin@slac.stanford.edu} \\
\And
Julia Gonski \\ 
SLAC National Accelerator Laboratory \\
Menlo Park, CA 94025 \\
\texttt{jgonski@slac.stanford.edu} \\
}

\begin{document}

\ifcolmsubmission
\linenumbers
\fi

\maketitle

\begin{abstract}
Recent advances in agentic AI have demonstrated the potential to automate complex high-energy physics (HEP) workflows. We focus on the regime of HEP tasks where frontier models still struggle to provide correct and complete solutions, and introduce an open-ended optimization task format with continuous, automatically evaluated rewards. Unlike conventional benchmarks focused on task completion, these tasks target reinforcement learning (RL) environments to generate agentic improvement on state-of-the-art HEP challenges. We benchmark frontier reasoning models on several HEP tasks and demonstrate an initial proof of concept in which a small language model improves through RL using automated feedback without human intervention. By connecting challenging HEP problems with verifiable scientific rewards and RL, we aim to advance agentic AI from reproducing established workflows to autonomously discovering better scientific solutions, paving the way toward self-improving scientific agents.

\end{abstract}

\section{Introduction}

Experiments at high energy particle colliders such as CERN's Large Hadron Collider (LHC) have significantly benefitted from modern advances in artificial intelligence and machine learning (AI/ML)~\cite{albertsson2019machinelearninghighenergy,duarte2025machinelearning,Radovic2018}.
The task of recording, reconstructing, and analyzing proton collision data from particle detectors like ATLAS~\cite{atlas_jinst} and CMS~\cite{Chatrchyan:1129810} requires complex feature extraction and inference over $\mathcal{O}(10^8)$ multi-modal inputs from several highly granular subdetectors.  
AI agents can autonomously design, optimize, and execute AI/ML workflows in HEP, allowing for high-performance data pipelines with dramatic reductions in scientific labor and time-to-result. 
Considerable recent work has explored the adoption of agentic tools to various key tasks in collider physics, such as physics analysis and discovery capability~\cite{gendreaudistler2025automatinghighenergyphysics,moreno2026aiagentsautonomouslyperform,he2026drsaiagenticairealworld,diefenbacher2026agentsdiscovery}, simulation and phenomenology~\cite{qiu2026endtoendarchitecturecolliderphysics,costa2026agentrivetautomatedproducingrivet,plehn2026madagents}, and to the design and operation of detectors and accelerators~\cite{hill2026graceagenticaiparticle,chung2026agenticaidetectorcodesign,hellert2026agenticartificialintelligencemultistage}. 

To fully meet their potential as autonomous scientific researchers, agentic systems must go beyond automating established workflows to support the development of new methods for open-ended research questions, while ensuring interpretability, verifiability, and reliable uncertainty quantification. 
Prior work has explored agentic harnesses guided by scientific requirements~\cite{liu2026scifisafelightweightuserfriendly} on close-ended tasks, but supporting robust, exploratory research remains an open challenge. 
Recent results from agentic researchers in fields with strong mathematical and theoretical requirements demonstrate the potential~\cite{openai-finite-time-blowup,turturean2026proofaaronsonambainisconjecture}.
However, experimental physics representing a fundamentally different set of tasks, requiring use and understanding of data and instruments, making this a key outstanding frontier for honing agentic tools. 

Reinforcement learning (RL) environments can be used as a model post-training approach where a model works within a closed system representing a task of interest, and learns by taking actions, getting feedback as rewards or penalties, and improving its strategy over many trials~\cite{Guo2025DeepSeekR1,ouyang2022traininglanguagemodelsfollow}. 
RL and related techniques like genetic algorithms can adapt general-purpose large language models (LLMs) to open-ended scientific reasoning tasks where they currently fall short. 
Ultimately, this can accelerate the development of agents that perform better on the kinds of open-ended, multi-expertise tasks that researchers actually face.

This work focuses on experimental collider research as a compelling setting for post-training and introducing the next generation of agentic researchers. 
We formulate several standard collider physics data analysis tasks for agentic systems and assess the limitations of frontier models on several representative tasks, including a newly designed task intended to probe their ability to develop analysis strategies beyond established templates. 
We introduce a task format consistent with an RL environment with continuous-valued rewards and evaluate it across these tasks, revealing a gap and/or incapability of several frontier models on the most challenging ones. 

To study the use of post-training to fill this gap, we investigate self-improvement by combining genetic-evolution and reinforcement learning post-training of a small open-weight model. 
We observe improvements in both execution reliability and the physics quality of the resulting solutions, including fewer crashes, more physically motivated and robust analysis strategies, and reduced reliance on reward exploitation or trivial solutions. Importantly, without additional domain-specific guidance, the improved model achieves performance comparable to out-of-the-box frontier models on held-out datasets for the same tasks. 
These results demonstrate the strong potential of task-specific self-improvement in small models and motivate exploring whether these gains can transfer across tasks, scale to more capable open-weight models, and support progress on open research questions in HEP and other scientific domains.

\section{Model Benchmarking on Collider Physics Research Tasks}
\label{sec:challenges}

For scientific applications of agentic tools, correctly executing an analysis workflow is necessary but insufficient.
Scientific value depends on sensitivity, precision, robustness, and correct intermediate reasoning stages, all meaning that ``correct" workflows can differ substantially in physics performance. 
Further, because research tasks are open-ended, binary success criteria is also insufficient; instead, evaluation must distinguish among correct solutions and reward meaningful improvements. 
We therefore require agents to submit executable scripts and score them with continuous, objective-aligned rewards on hidden datasets, measuring both reusability and performance gains beyond baseline methods.

With these considerations in mind, we define physics-motivated rewards and use them to evaluate several frontier and open-weight models on six collider physics tasks.
Each task is specified by a physics objective, development data, a deliverable, and a scoring rule, leaving agents free to develop their methods. 
We provide a minimal execution interface without a domain-specific harness, intended to assess the models' out-of-the-box ability to develop physics analysis methods with minimal external scaffolding. Each model autonomously writes, executes, and revises its scripts within a single session, guided only by the task description, development data, and feedback from its own analysis and executions. 
To assess reusability and physics performance, we execute submitted scripts on hidden datasets in a fresh working directory under a fixed CPU budget. Hidden-data scores are used solely for evaluation and are not returned to the agents.

\paragraph{Physics-driven reward functions.} We align each reward with the physics goal through the corresponding metric such as discovery significance, signal sensitivity, reconstruction accuracy, background rejection or distribution fidelity.
We reward physics performance directly, without normalizing it to literature results or in-house baselines, which serve only as reference points. 
This avoids embedding prior expectations about achievable performance in the reward and leaves room for unrestricted methodological exploration. 
Since there is no single correct solution, the best-known human result does not impose an upper bound on the reward.

\paragraph{Continuous scoring.} We quantify partial progress by mapping the metric to $R\in[0,1]$ with a monotonic, saturating function that also provides a learning signal. For example, in the task \texttt{HiggsML}, which is detailed in Sec~\ref{HiggsML}, the reward $R = \max\!\big(0,\,1-\exp(-(\mathrm{AMS}-2.5))\big)$ uses the approximate median significance of the 2014 challenge~\cite{adambourdarios2015higgsml}: the untuned benchmark earns 0.50 and the winning entry 0.73, while improved selections earn higher rewards.

\paragraph{Multi-objective construction.} We combine critical physics criteria multiplicatively, so that failure on a required criterion gives zero reward to reject cheating or non-physical solution. For the anomaly detection task in Sec~\ref{AD}, the reward $R = S^{3/4}\,Q^{1/4}$ balances signal sensitivity $S$ against mass-sculpting control $Q$, while unfolding and fast simulation tasks in Sec~\ref{unfold} and \ref{calo}, combine distribution metrics with guards against shortcuts. Tracking task in Sec~\ref{track} uses the official score in the publication; across tasks, missing, invalid, crashed or timed-out submissions score zero.

\paragraph{Model evaluation} To assess the out-of-the-box behavior of models under this reward framework, we adopt a simplified, common autonomous protocol with no prescribed physics workflow. 
Each model receives the task description and access to a shell for a single 30-minute session. We score submissions on hidden data to obtain reward trajectories, without exposing these hidden scores to the agents. Unless otherwise indicated, we report results for the final submissions, with trajectories shown in Figures~\ref{fig:outofbox}-\ref{fig:casestudy_benchmark}. 
Transcript review confirmed that the models used only permitted inputs. 
We evaluate Claude Opus~5.5, Claude Sonnet~5.5, Claude Haiku~5.5, GPT-6.1-sol, GPT-6-luna, Gemini~4 Argon, Grok~4.7, DeepSeek~V4.1 Flash, GLM-5.3 Prime, and Qwen3.8-Max Prime; not every model is evaluated on every task due to resources budget.

Five tasks are drawn from public challenges that are well documented in the literature and may be represented in LLM training data (Sections~\ref{HiggsML}-~\ref{track}). 
The sixth is a newly designed model-agnostic diphoton peak search, intended to probe zero-shot performance and examined in detail in Section~\ref{sec:casestudy}.
\textbf{This evaluation reveals two distinct patterns: convergence toward human-level performance on well-established ``easy" tasks, and substantial variation in performance on zero-shot state-of-the-art tasks.}
Details of each evaluation are discussed in the following sections. 

\subsection{HiggsML}
\label{HiggsML}

\begin{figure}[t]
\begin{center}
\includegraphics[width=0.75\linewidth]{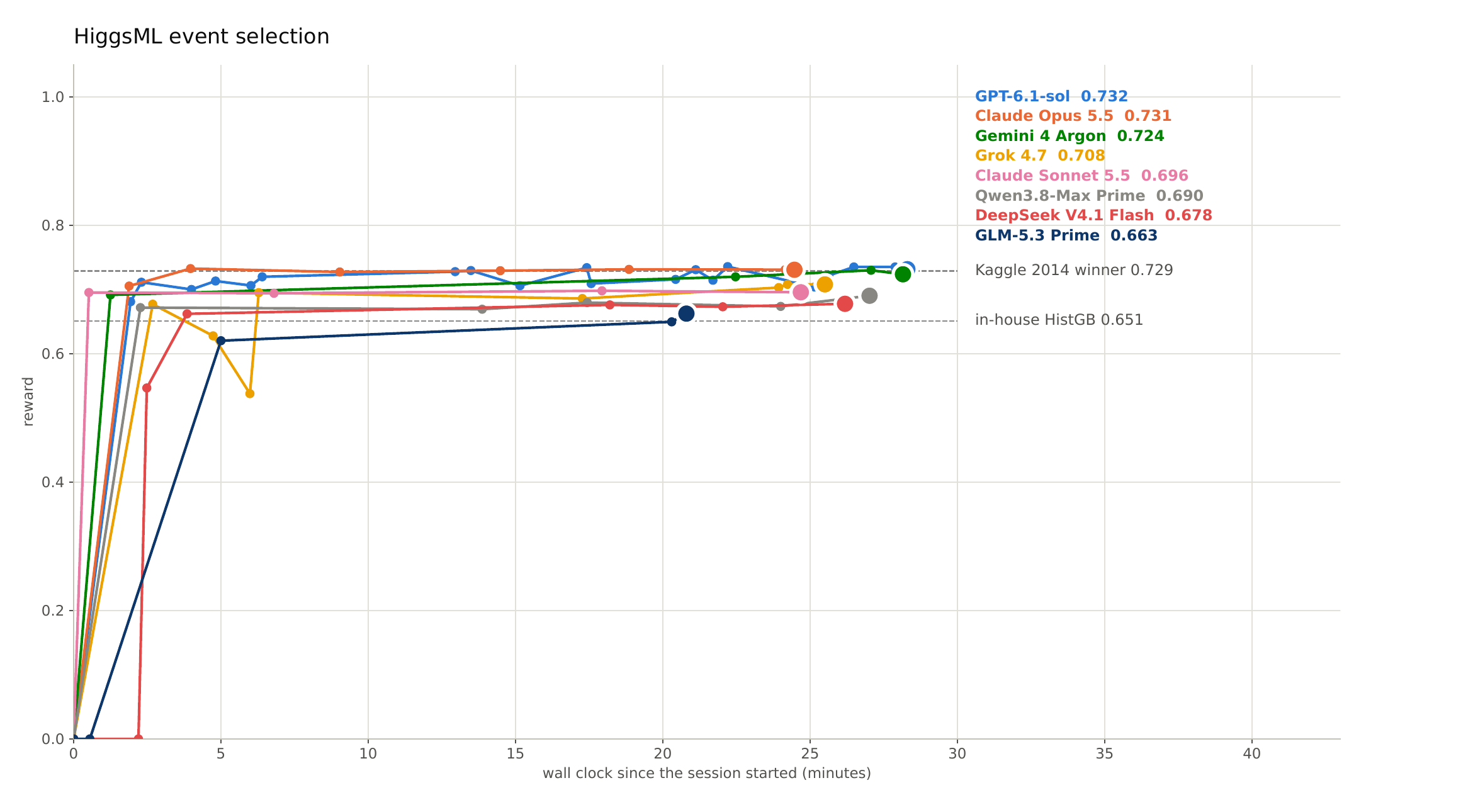}
\end{center}
\caption{Reward over time on HiggsML task. Dashed references are the Kaggle 2014 winner and in-house baseline solution using gradient boosting model.}
\label{fig:outofbox}
\end{figure}

HiggsML~\cite{adambourdarios2015higgsml} evaluates event selection over simulated datasets to identify a $H\to\tau\tau$ signal with respect to background as a binary classification problem.  
Agents select from the supplied events, and we score their selection by its approximate median significance (AMS) to measure search sensitivity.
Figure~\ref{fig:outofbox} shows the reward as a function of time spent on the problem for the HiggsML task. 
All models considered arrive at human-level performance, and all use the same approach: boosted trees with physics features and reach 0.708--0.732, near the 2014 Kaggle challenge winner (0.729).
We conclude that simple event classification in collider physics is easily achieved within the existing capabilities of both frontier and other models. 

\subsection{Unfolding}
\label{unfold}

Unfolding is the procedure of inferring the underlying truth (particle-level) distributions from detector-level observations by correcting for detector resolution, inefficiencies, and acceptance effects.
The unfolding task~\cite{huetsch2024landscapeunfoldingmachinelearning} tests the recovery of particle-level distributions on the OmniFold $Z$+jets benchmark~\cite{Andreassen2020OmniFold}. 
Given paired particle- and detector-level simulation and detector-level data from a different generator, agents deliver event weights. 
We assess agreement with hidden particle-level distributions in six substructure observables and their correlations.

\begin{figure}[t]
\begin{center}
\includegraphics[width=0.75\linewidth]{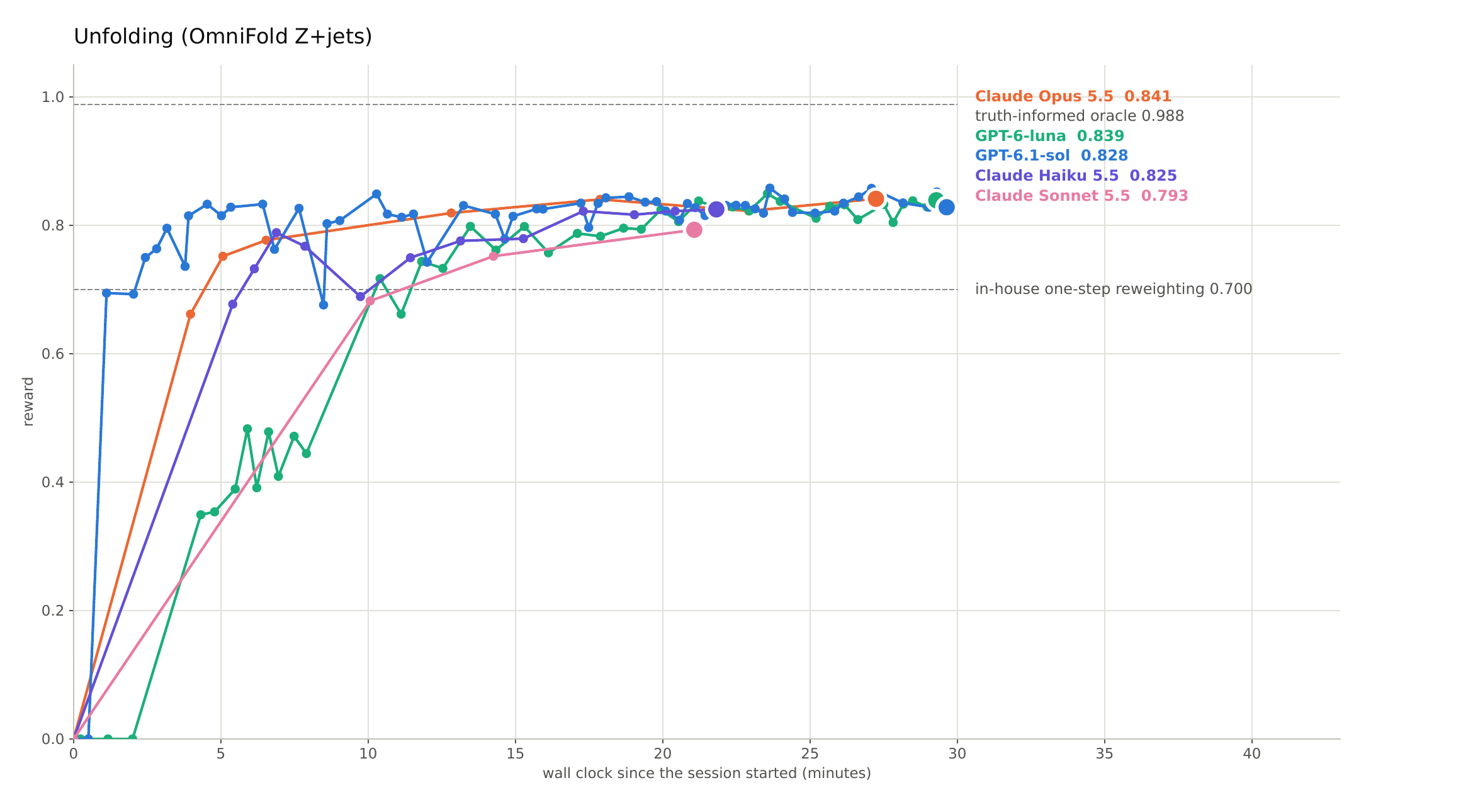}
\end{center}
\caption{Reward over time on unfolding task. Dashed references are a truth-informed oracle and in-house baseline solution using one-step reweighting.}
\label{fig:outofbox_unfolding}
\end{figure}

As shown in Figure~\ref{fig:outofbox_unfolding}, all five models converge on iterated OmniFold reweighting method, independently implementing the method and reaching 0.793--0.841, above the one-step baseline (0.700). 
However, none reach the truth-informed oracle solution. 
We conclude that unfolding is doable by LLM tools to best-achieved human-level performance. 

\subsection{LHC Olympics anomaly detection}
\label{AD}

LHC Olympics~\cite{kasieczka2021lhcolympics2020community} tests the identification of an unknown resonance in dijet events using a supplied QCD background sample. Agents assign anomaly scores, with a reward that balances signal sensitivity against background sculpting. A resonance is a new particle visible as a localized excess in the invariant-mass spectrum of its decay products. 
Sculpting occurs when the selection itself creates a bump in the smooth background spectrum, resulting in possible false positives or indication of a signal where there is none. 

\begin{figure}[t]
\begin{center}
\includegraphics[width=0.75\linewidth]{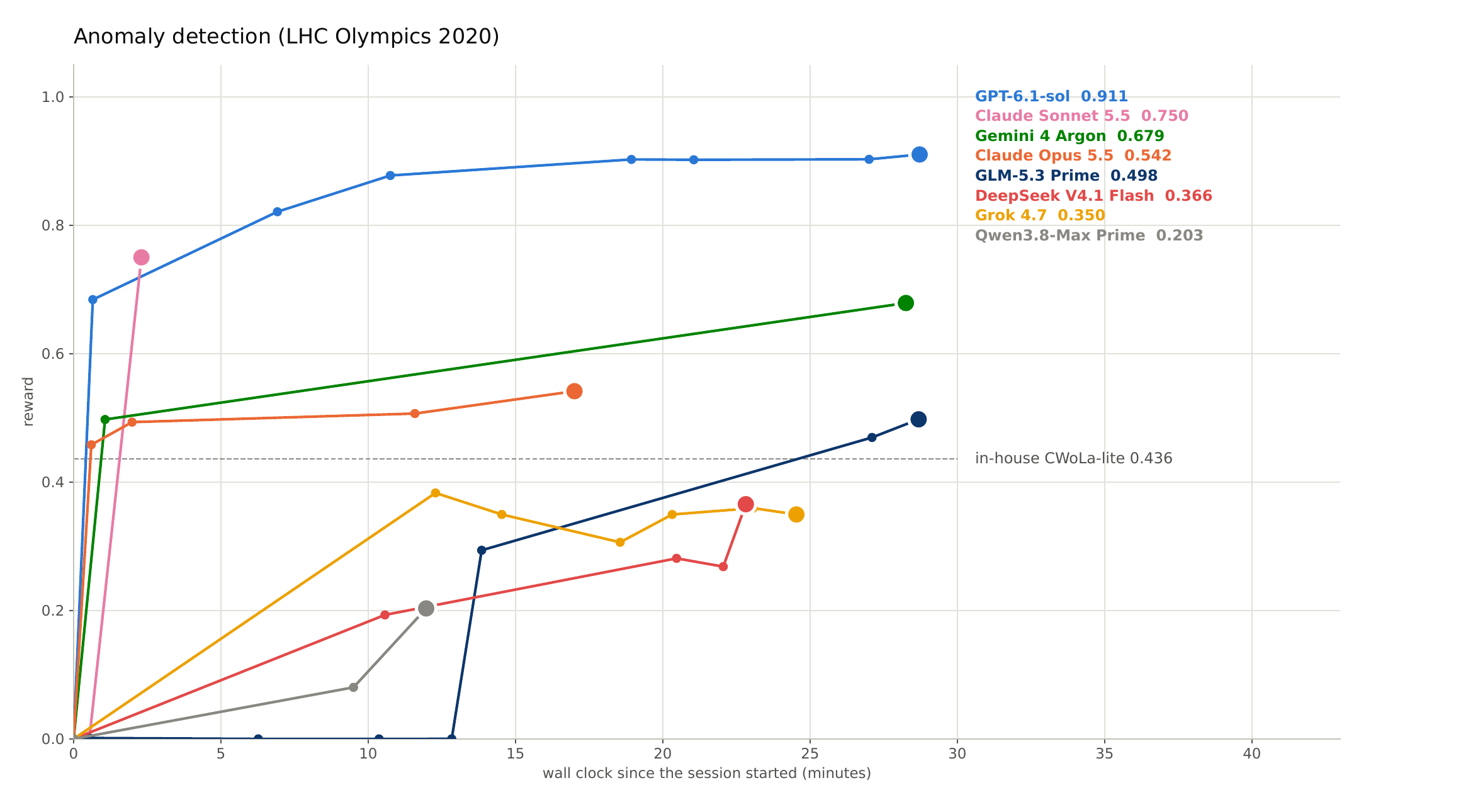}
\end{center}
\caption{Reward over time on LHC Olympics anomaly detection task. The dashed reference is an in-house baseline solution using CWoLa-lite\cite{metodiev2017cwola,collins2018cwolahunting} classifier.}
\label{fig:outofbox_anomaly}
\end{figure}

Figure~\ref{fig:outofbox_anomaly} shows the model performance over time for the anomaly detection task, revealing substantial differences in validation strategy and performance, with scores across eight models ranging from 0.203 to 0.911. 
The best-performing session trains against the supplied QCD background in the development signal region and validates on weak-signal and signal-free samples, approximating the background conditions in the test data. Other sessions train classifiers using strictly sideband data but choose suboptimal thresholds based on the development data; their estimated scores exceed the hidden-data scores by up to 0.55. Three models score below the in-house CWoLa baseline (0.436), indicating difficulty in implementing or developing effective methods. It worth noting that even the highest-scoring solution has an important limitation: it relies on the supplied development sample to model the background in the signal region, effectively treating it as a reliable Monte Carlo prediction. Its strong benchmark performance therefore does not establish robustness in broader applications, where mismodeling of the simulated background can compromise the analysis. This limitation also motivates the zero-shot task introduced in Section~\ref{sec:casestudy}, which considers anomaly detection on other final states with deliberately less accurate Monte Carlo background modeling.

\subsection{CaloChallenge}
\label{calo}

CaloChallenge~\cite{krause2025calochallenge2022communitychallenge} task tests generative modelling of Geant4 photon showers in a voxelised calorimeter. 
Agents construct a surrogate from supplied showers; we assess its distribution fidelity on hidden data, with guards against shortcuts.

\begin{figure}[t]
\begin{center}
\includegraphics[width=0.75\linewidth]{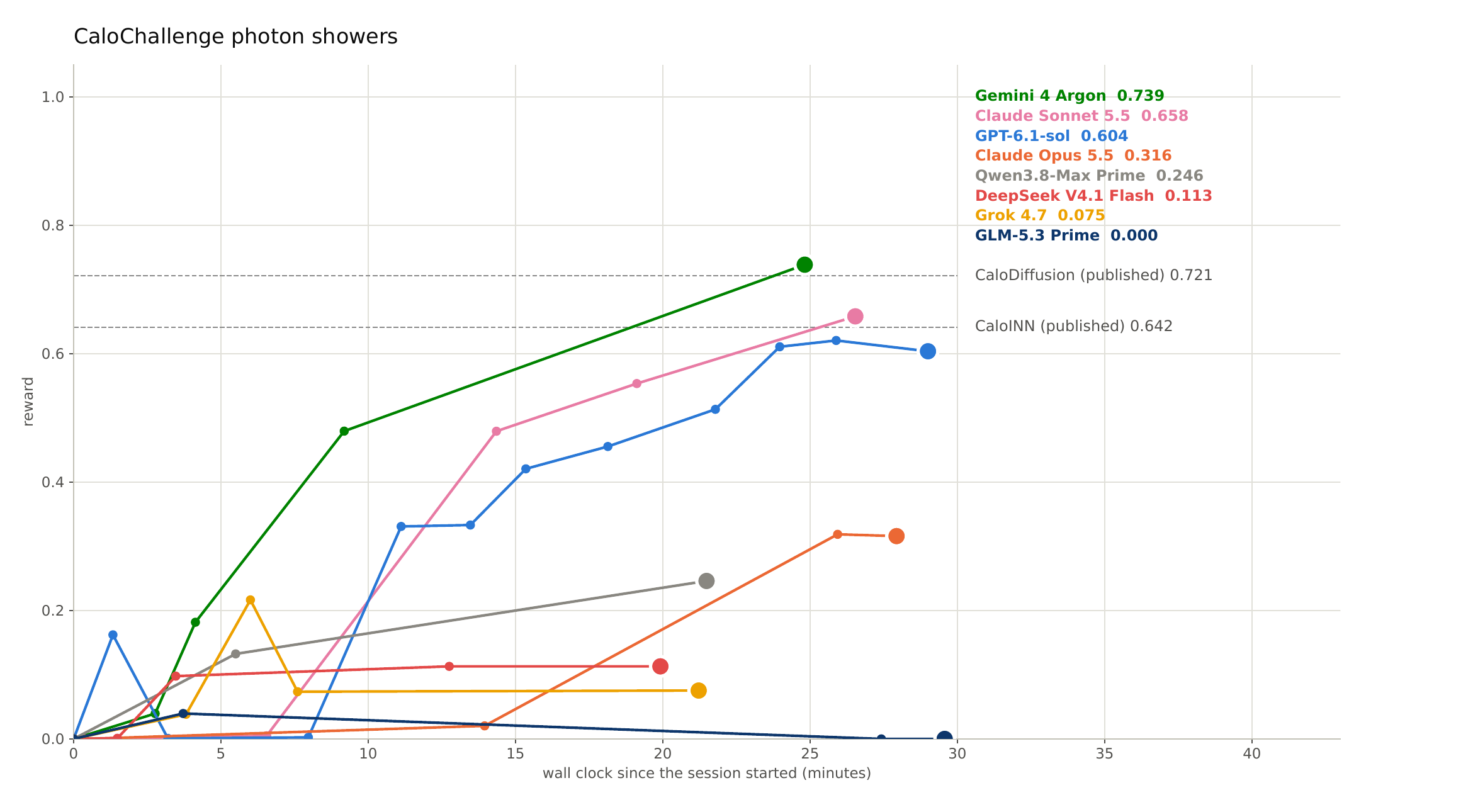}
\end{center}
\caption{Score over time per model on CaloChallenge task; dashed lines are published solution scored with the reward-based task metric.}
\label{fig:outofbox_open}
\end{figure}

Model performance as shown in Figure~\ref{fig:outofbox_open} varies widely on the CaloChallenge task, with final rewards ranging from 0.000 to 0.739. 
Gemini 4 Argon (0.739) and Claude Sonnet 5.5 (0.658) achieve performance comparable to or above that of published generative models (0.64--0.72), while GPT-6.1-sol scores slightly below this range (0.603). 
The two best-performing sessions use per-energy Gaussian mixtures in a rank-Gaussianised voxel space (Gemini) and a small conditional diffusion model (Sonnet), respectively, both followed by quantile calibration to match the training marginals. 
By contrast, the remaining five models score below 0.32 and rely on Gaussian copulas that inadequately capture the 368-dimensional correlation structure, allowing the evaluation classifier to readily distinguish generated showers from reference samples.
The  high-dimensional and generative nature of this task reveal inconsistencies among frontier models, as well as room for improvement even in the state-of-the-art. 

\subsection{TrackML}
\label{track}

The TrackML task~\cite{amrouche2021trackingmachinelearningchallenge} is based on a simplified LHC tracking scenario in which each collision produces roughly 10,000 charged-particle tracks (approximately helicoidal trajectories emerging from near the interaction point) that leave about 10 high-precision 3D hits each. 
The core pattern recognition task is to associate the resulting ~100,000 space points into their corresponding tracks, a problem where traditional tracking approaches can suffer a combinatorial explosion in CPU time as occupancy increases.
Agents group the supplied hits into tracks, which we assess with the official double-majority score.

\begin{figure}[t]
\begin{center}
\includegraphics[width=0.75\linewidth]{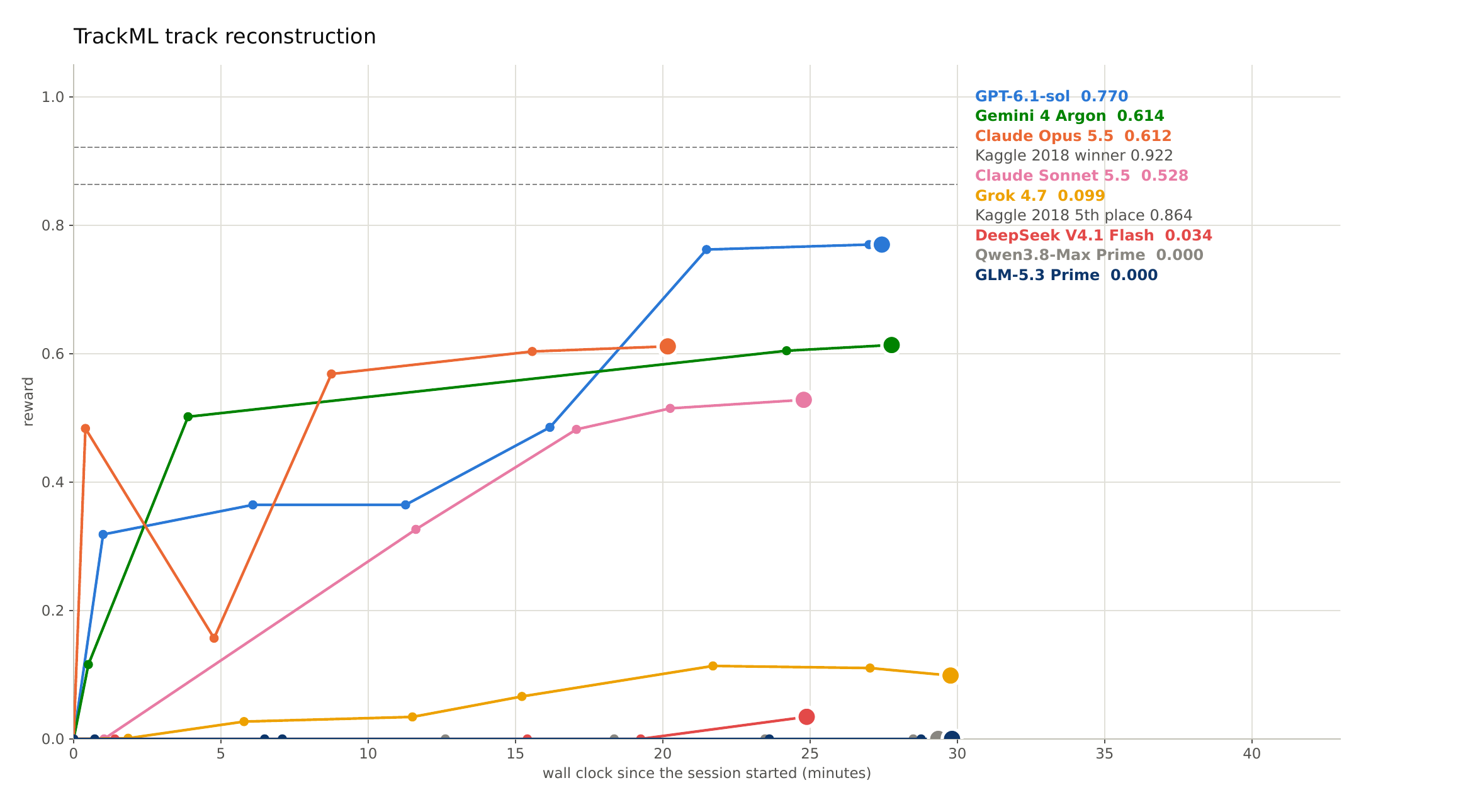}
\end{center}
\caption{Score over time per model on TrackML; dashed lines are the 2018 leaderboard scored with the reward-based task metric.}
\label{fig:outofbox_trackml}
\end{figure}

This task also represents a genuine challenge and feasibility frontier for the models considered, as shown in Figure~\ref{fig:outofbox_trackml}.
Track reconstruction remains below published performance across models, with final rewards ranging from 0.000 to 0.770. The best-performing session, GPT-6.1-sol (0.770), builds a seeding-and-road-following pipeline with a learned candidate-quality model, structurally similar to the 2018 winning solutions (0.86--0.92) but substantially less refined. Gemini, Opus, and Sonnet (0.53--0.61) converge on a simpler approach used in public competition kernels: clustering hits after unrolling candidate helices over a scan of curvature values. The remaining four models exhaust the time budget before producing functional tracking algorithm and receive negligible rewards.

\subsection{Diphoton resonance search}
\label{sec:casestudy}

The final task is developed explicitly for this study and is discussed in-depth, as it is subsequently used for a post-training proof-of-concept.  
The resonance search task asks agents to construct a reusable search that identifies zero, one or two narrow resonances and reconstructs their masses. 
We provide full event kinematics for simulated proton--proton collisions with two photons and up to three jets in $m_{\gamma\gamma} = 105$--$145$\,GeV. 
Any method is allowed, including anomaly detection style weakly supervised classifiers or direct background estimation from each dataset's sidebands, provided that programs read boundaries, luminosity and selections from the dataset's metadata. 

\begin{figure}[t]
\begin{center}
\includegraphics[width=0.99\linewidth]{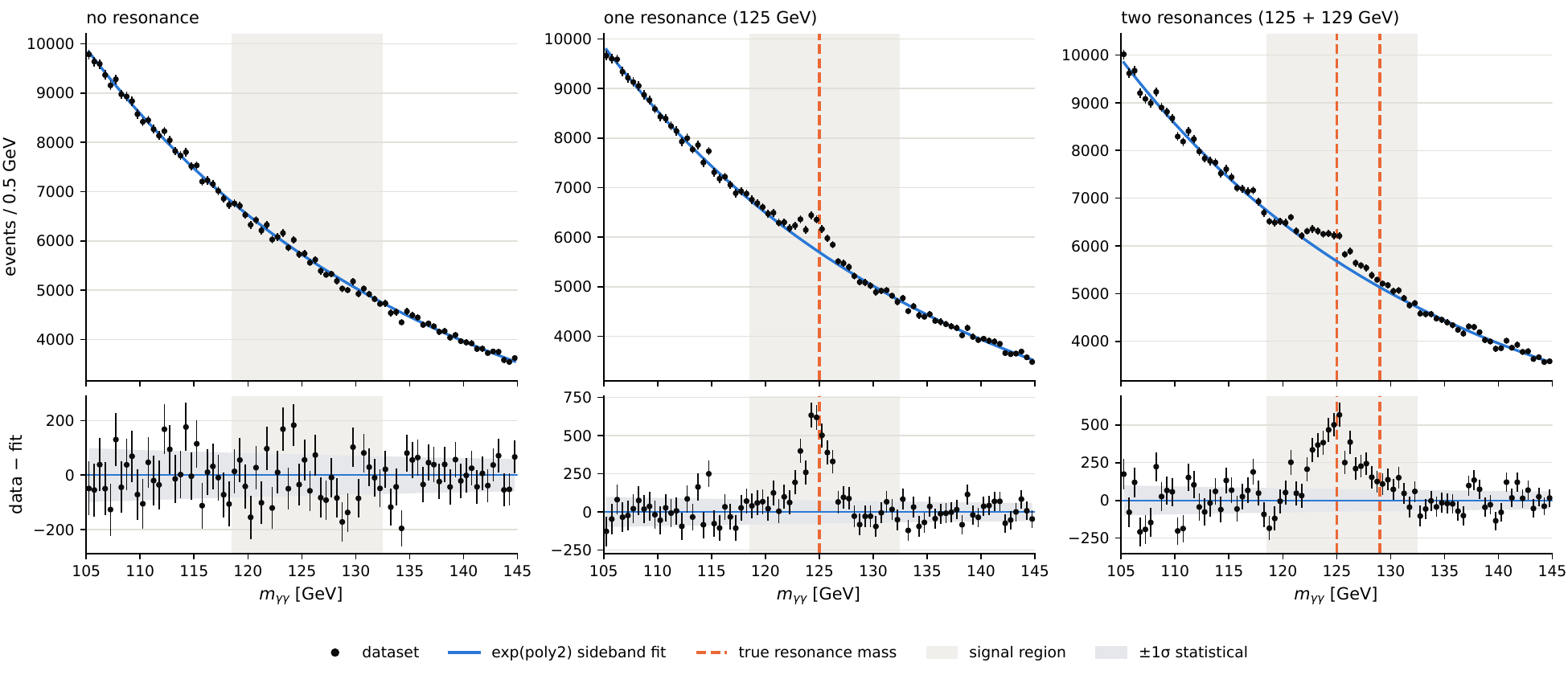}
\end{center}
\caption{Three of the twelve hidden datasets: no resonance; one at 125\,GeV; and two at 125\,GeV and 129\,GeV. Top: diphoton mass spectra, with the signal region 118.5--132.5\,GeV shaded and an illustrative exponential-quadratic sideband fit. Bottom: residuals from that fit with a $\pm1\sigma$ statistical band. Dashed lines mark true masses (hidden information).}
\label{fig:casestudy_data}
\end{figure}

The development sample supplies about 800\,000 events, ground truth, weighted signal and background templates, and an auxiliary continuum sample, intended to be a low-precision simulation. 
The twelve hidden datasets use a different simulation setup, with four datasets per resonance count (Fig.~\ref{fig:casestudy_data}). 
The datasets combine continuum background events generated with MadGraph5 interfaced with Pythia8, gluon-fusion Higgs and heavy-vector-triplet (HVT) signals, and detector simulation using a custom Delphes card based on the ATLAS geometry, with configurations chosen to avoid directly reproducing published results. 
Each dataset contains approximately 159\,000 continuum background events in the signal region of 118.5--132.5\,GeV, alongside injected resonances with widths of 1.35--1.50\,GeV. The 125\,GeV Higgs resonance contributes approximately 3\,100 signal events, corresponding to a local significance of $S/\sqrt{B}=11.7$ within a $\pm 2\sigma$ window. The HVT resonances contribute 380--1\,370 events, with local significances of 1.4--5.5, placing most hidden signals near or below the conventional discovery threshold.

This task is designed to encompass several key challenges. 
The search is complicated by weak signals, overlapping resonances, and uncertainties in background estimation. The non-Higgs signals have expected local significances of $S/\sqrt{B}=1.4$--$5.5$ within a $\pm2\sigma$ mass window. Background estimation relies on fitting the sidebands and interpolating into the signal region, introducing uncertainties that can substantially alter the apparent significance of a resonance. For example, a nominal $3\sigma$ excess at 121\,GeV is reduced to $1.2\sigma$ after fitting, whereas a $1.4\sigma$ excess at 124\,GeV increases to $2.6\sigma$. The resonance widths range from $\sigma=1.35$ to $1.50$\,GeV, with two pairs separated by only 2\,GeV, further complicating their resolution. Additionally, the 125\,GeV Higgs resonance is present in only five of the eight datasets containing injected signals, precluding any assumption of its presence. This setting differs from conventional diphoton analyses and existing publications, which typically target a known Higgs mass and employ predefined signal templates and they are not generally optimal in this task.

The reward is designed to measure both resonance counting and mass reconstruction. 
We match claimed masses to true resonances one-to-one to maximise total credit, with each matched pair earning
\begin{equation}
c(\hat m, m) = \max\!\left(0,\; \frac{e^{-(\hat m-m)^2/2s^2} - e^{-w^2/2s^2}}{1 - e^{-w^2/2s^2}}\right), \qquad s = 0.5\,\mathrm{GeV},\; w = 1\,\mathrm{GeV},
\end{equation}
which decreases with mass error and vanishes beyond $w$. For $n_{\mathrm{true}}$ true and $n_{\mathrm{claimed}}$ claimed peaks, the soft F1 score $2\sum c / (n_{\mathrm{true}} + n_{\mathrm{claimed}})$ rewards the correct count and accurate masses while penalising false claims and missed signals. A background-only dataset scores 1 for no claim and 0 otherwise. All the scores are averaged across all datasets. Consequently, always claiming no resonance scores 0.333, while always claiming 125\,GeV scores 0.250. Our reference program fits an exponential-quadratic background to the sidebands, scans a Gaussian template, subtracts the leading excess and scans again. It achieves 0.689 overall, with 1.000, 0.599 and 0.468 across the resonance-count groups.

\begin{figure}[t]
\begin{center}
\includegraphics[width=0.8\linewidth]{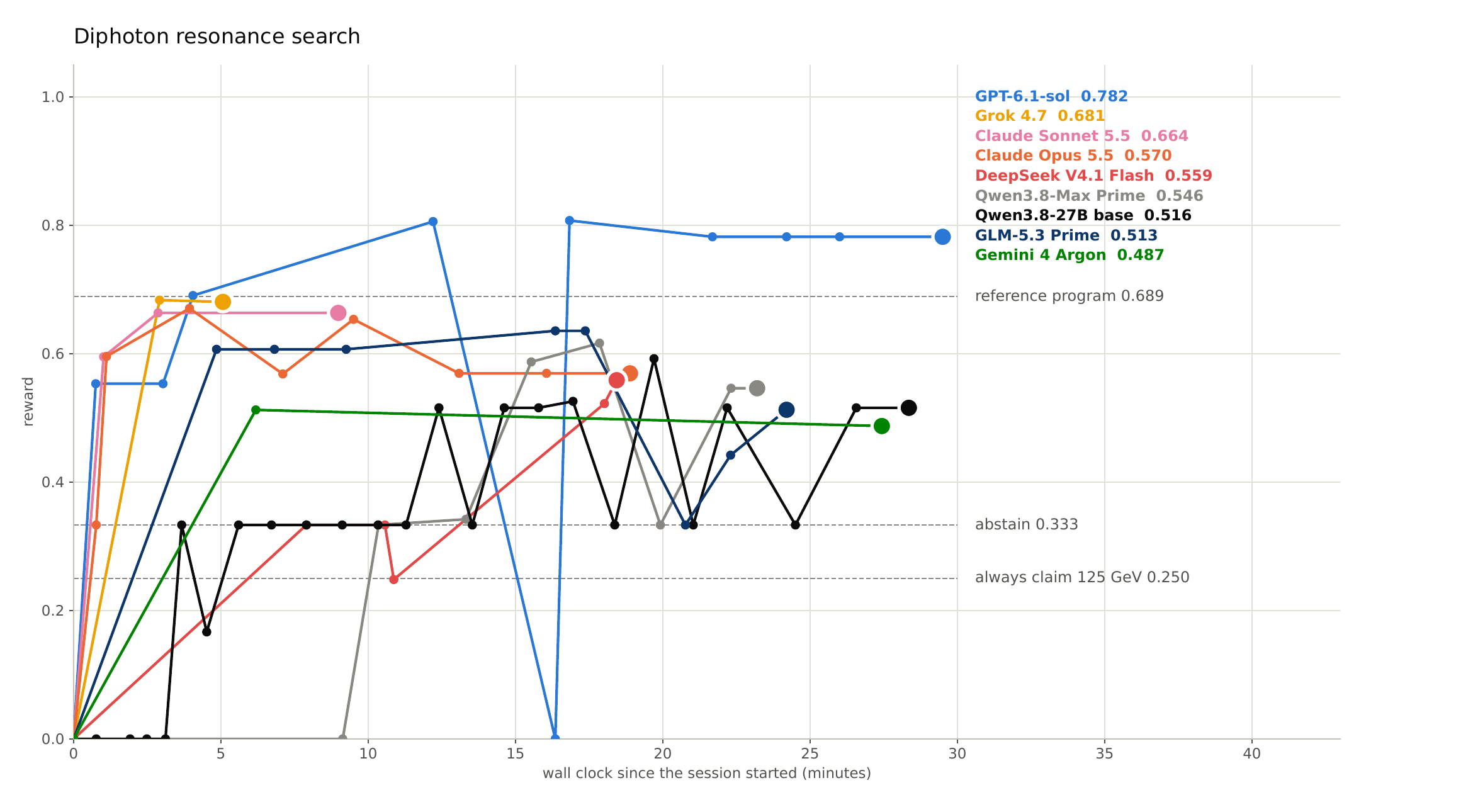}
\end{center}
\caption{Reward over time on the twelve hidden datasets for the diphoton peak search task. Black shows Qwen3.8-27B, the base model in Section~\ref{sec:selfimprove}, with execution feedback. Dashed lines show the reference program (0.689), always abstaining (0.333), and always claiming 125\,GeV (0.250).}
\label{fig:casestudy_benchmark}
\end{figure}

Figure~\ref{fig:casestudy_benchmark} shows the reward as a function of time spent for several frontier models. No model saturates the reward; final rewards range from 0.487 to 0.782, and only GPT-6.1-sol (0.782) exceeds the reference. 
The open-weight Qwen3.8-27B, the only non-frontier model in this comparison, achieves a reward of 0.516 in the standard 30 minute session. 
To facilitate a more meaningful comparison with frontier models, we additionally provide it with reward feedback evaluated on an independent development dataset, separate from the held-out dataset used for final reporting. 
This additional feedback enables the model to iteratively refine its solution and narrow the performance gap with frontier models, although a noticeable gap remains.

Notably, all sessions produce valid programs with background models, Gaussian-excess scans and significance thresholds. 
However, the actual correctness of the achieved solution varies considerably: six of the eight programs make no false claim on the background-only datasets, yet all lose reward on the one-resonance group (0.214--0.730) and especially the two-resonance group (0.235--0.616).
We identify four failure modes:

\begin{enumerate}
\item \emph{Sequential searches miss overlapping peaks.} Most programs search for the strongest peak before examining the residual, sometimes imposing a minimum separation of 2--2.5\,GeV. Consequently, five of the eight never report a second mass on any two-resonance dataset, while the others usually place it more than 1\,GeV from the true mass.

\item \emph{Aggressive cuts produce false discoveries.} One session lowers thresholds into the tails of its background-only trials to recover sensitivity to pairs. Although the final program reports two masses on every two-resonance dataset, it also makes a false background-only claim, exposing the sensitivity--purity trade-off.

\item \emph{Limited exploration.} Three sessions stop early despite known weaknesses. Several programs also retain development widths or background templates despite the broader hidden peaks; GPT-6.1-sol instead fits widths freely within 0.65--2.3\,GeV and performs best.

\item \emph{Technical failures.} Two sessions overwrite working submissions with broken programs before repairing them. In two sessions, histogramming or fitting bugs delay the first valid submission until 8--13 minutes.
\end{enumerate}

The transcripts therefore point to limitations in exploration and self-verification under the physics objective, even though the methods, toy-based calibration and two-peak logic already appear. This motivates the study in Section~\ref{sec:selfimprove}, where we test whether the model can self-improve its search from the reward alone.
Table~\ref{tab:outofbox} provides a summary of all six tasks, the model reward range, and the human/oracle reference used. 

\begin{table}[t]
\centering
\small
\resizebox{\linewidth}{!}{%
\begin{tabular}{@{}lccp{7.2cm}@{}}
\toprule
Task & Models & Reward range & Reference \\
\midrule

HiggsML event selection
& 4 & 0.708--0.732
& HiggsML winner~\cite{Melis2015HiggsML}
  (0.729, rescored);
  in-house HistGB (0.651) \\

Unfolding (OmniFold $Z$+jets)
& 5 & 0.793--0.841
& OmniFold~\cite{Andreassen2020OmniFold};
  in-house one-step baseline (0.700);
  truth-informed oracle (0.988) \\

Anomaly detection (LHCO 2020)
& 8 & 0.203--0.911
& LHCO~\cite{Kasieczka2021LHCO};
  in-house CWoLa-lite~\cite{Collins2018CWoLa}
  (0.436) \\

CaloChallenge photon showers
& 8 & 0.000--0.739
& CaloDiffusion~\cite{Amram2023CaloDiffusion}
  (0.72); CaloINN~\cite{Ernst2025CaloINN}
  (0.64); both rescored using our metric
  ~\cite{Krause2025CaloChallenge} \\

TrackML track reconstruction
& 8 & 0.000--0.770
& TrackML~\cite{Kiehn2019TrackML};
  Kaggle winner (0.922), fifth place (0.864)
  ~\cite{TrackML2018Leaderboard} \\

Diphoton resonance search)
& 8 & 0.487--0.782
& In-house reference program (0.689) \\

\bottomrule
\end{tabular}}
\caption{Out-of-the-box performance of frontier models
across six particle physics tasks.
Each model is evaluated in a single 30-minute session per
task, with the final submitted script scored on hidden data.
Reward ranges summarize performance across the evaluated
models. Reference scores provide task-specific benchmarks;
values labeled as rescored are evaluated using our reward metric.}
\label{tab:outofbox}
\end{table}

\section{Self-Improvement of LLM Models on Collider Physics Tasks}
\label{sec:selfimprove}

We combine genetic evolution and reinforcement learning to improve the diphoton resonance search task in Section~\ref{sec:casestudy} without prescribing a solution method. The procedure uses only the task, its public development data and reward, with no human-written solution or harness encoding a physics workflow. 
We use Qwen3-4B-Instruct-2507 (4B) and Qwen3.8-27B (27B), with base weights and thinking disabled, to explore the potential of the proposed approach for future frontier-level open-weight models. This study focuses on improving the models' intrinsic capabilities rather than enhancing performance through agentic workflows or external harnesses, which are beyond its scope. Complementary approaches based on agentic workflows using harness are discussed in Refs.~\cite{liu2026scifisafelightweightuserfriendly,gendreaudistler2025automatinghighenergyphysics}.

Optimization is performed using synthetic datasets constructed from the development sample, while final performance is evaluated on the twelve hidden datasets described in Section~\ref{sec:casestudy}, which are strictly excluded from optimization. Detailed experimental settings and prompts are provided in Appendix Table~\ref{tab:settings} and Appendix Section~\ref{app:prompts}, respectively.

\subsection{Genetic algorithms}
\label{sec:genetic}

\begin{figure}[t]
\begin{center}
\includegraphics[width=0.99\linewidth]{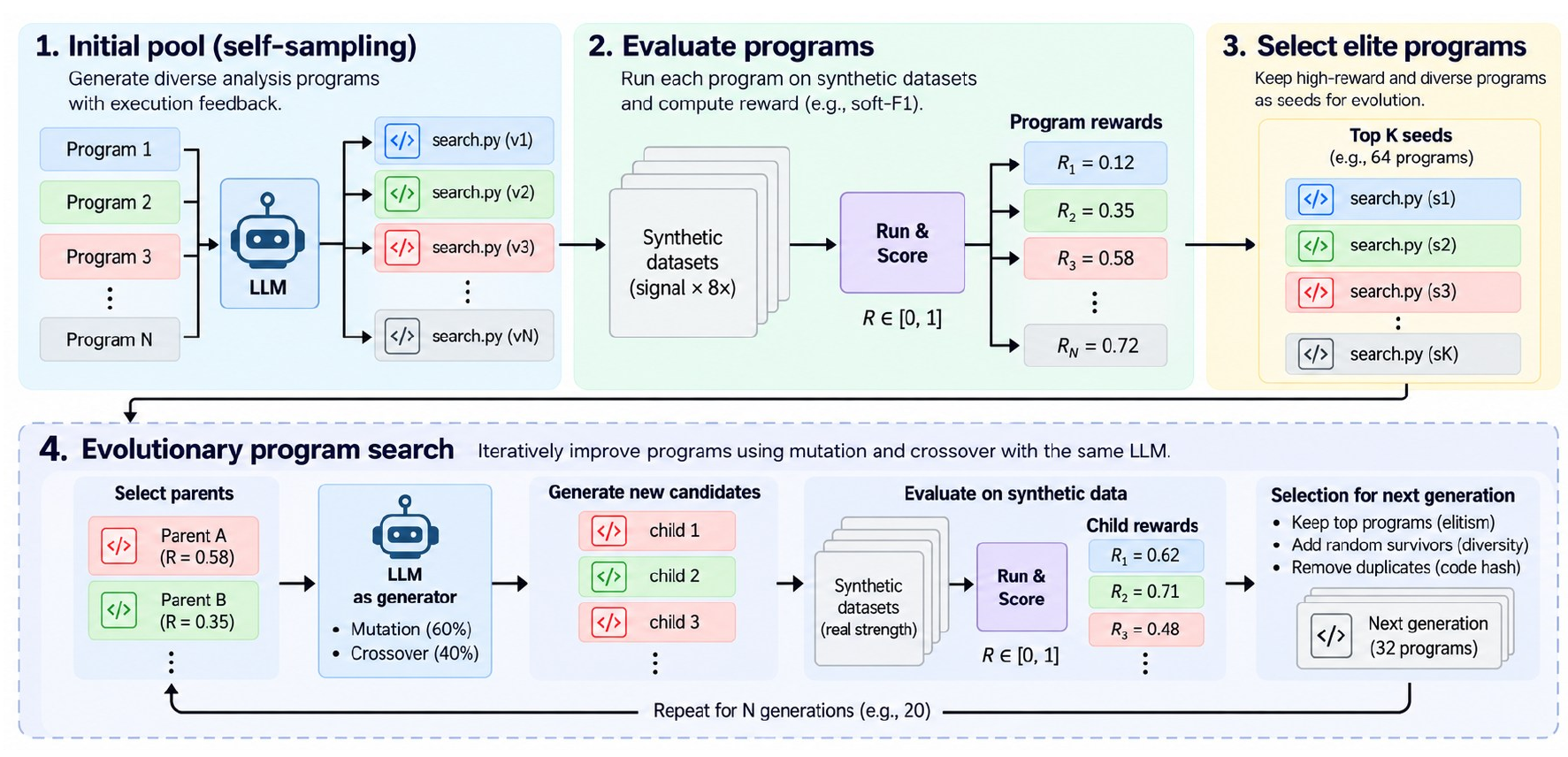}
\end{center}
\caption{Evolution with a frozen LLM. Self-sampling with execution feedback builds a program pool. The same model then proposes mutations and crossovers, ranked by the task reward on synthetic datasets.}
\label{fig:genetic}
\end{figure}

Sampling complete programs from a frozen LLM exposes substantial variation in both execution reliability and physics performance (Fig.~\ref{fig:genetic}). We execute and score each program, finding that rewards depend strongly on the sampled analysis and model size: some programs fail technically, while others run but do not improve the physics score. For the 27B model, the 32 initial programs ("base") average 0.116 on the twelve hidden datasets; 15 crash, 3 contain no code, and only one exceeds the trivial baseline, reaching 0.578. The 4B model averages 0.013 with no program above the baseline, motivating selection and revision rather than sampling alone.

We guide evolution by ranking programs on synthetic reward and returning selected programs and execution results to the same model for analysis and revision. Self-sampling with execution feedback first constructs a pool of candidate solutions using synthetic datasets with stronger injected signals, allowing the model to bootstrap its exploration from simple search strategies. This procedure does not require domain-specific knowledge, as the same approach can be applied in general. We test revised programs on fresh synthetic data to prevent reuse of revealed resonance masses, then evolve a population of 32 over 20 generations through mutation and crossover. Mutation revises a selected parent using its execution results, while crossover combines selected parents; fitness is measured on two fixed synthetic variants at real signal strength. High-fitness programs survive alongside random survivors to preserve exploration. Appendix~\ref{app:prompts} documents the prompts for both stages.

\begin{figure}[t]
\begin{center}
\includegraphics[width=0.9\linewidth]{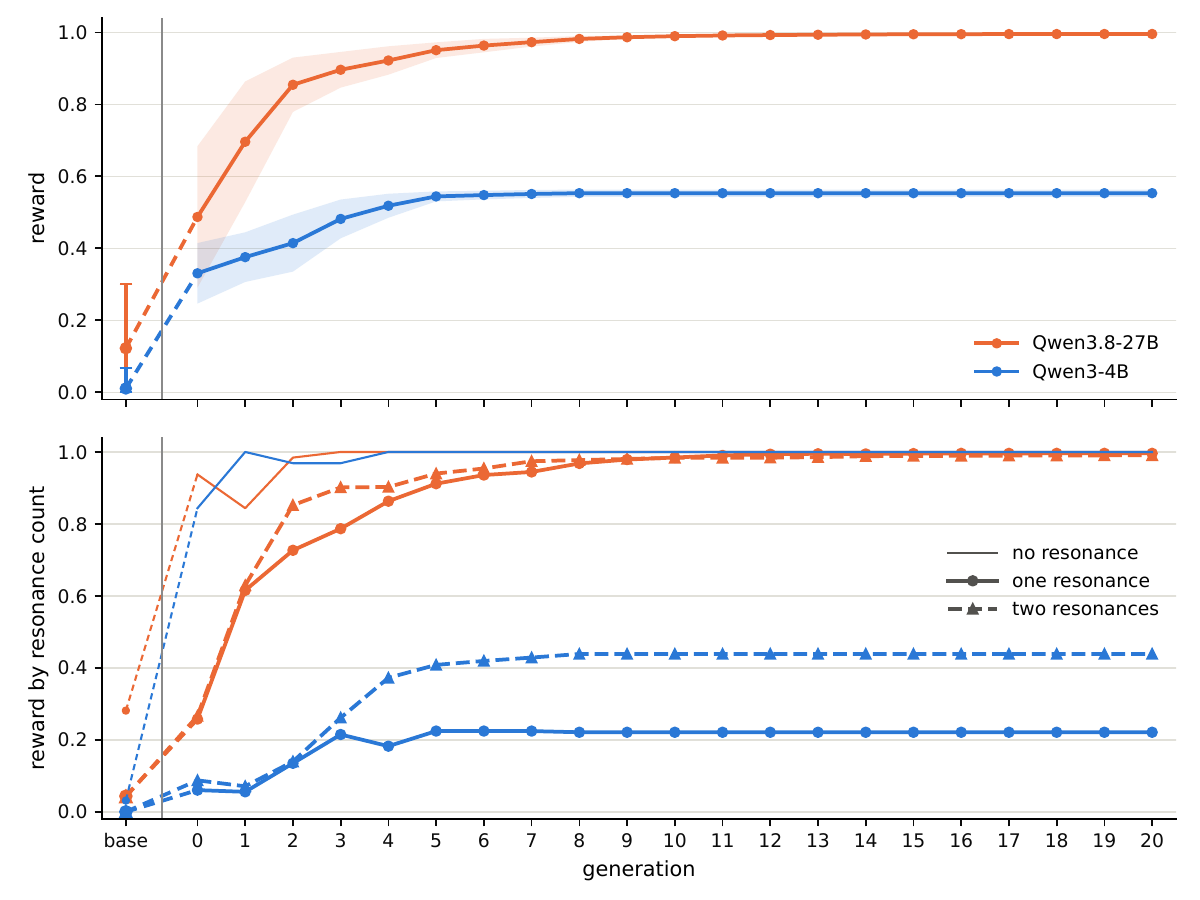}
\end{center}
\caption{Self-evolution of Qwen3.8-27B (orange) and Qwen3-4B (blue) on the synthetic fitness datasets. Top: reward of the 32 programs selected at each generation (mean and standard deviation). Bottom: their mean score on the synthetic datasets with no, one and two resonances. The rewards of the same programs on the twelve hidden datasets are given in Table~\ref{tab:results}.}
\label{fig:steps27b}
\end{figure}

Figure~\ref{fig:steps27b} shows two assessments of the task reward as a function of generation, revealing that evolution improves both synthetic fitness and performance on the hidden datasets. 
The 27B population's synthetic reward rises from 0.49 at generation 0 to 0.99 from generation 10, while its mean on the twelve hidden datasets increases from 0.382 at generation 0 to 0.626 at generation 12 (Table~\ref{tab:results}). 
Selecting by highest synthetic fitness yields a program scoring 0.644 [0.49, 0.76], an improvement over the base single-shot mean of $+0.51$ [$+0.37$, $+0.62$]. 
These 95\,\% paired-bootstrap intervals resample the twelve datasets within resonance-count groups. 

\begin{figure}[t]
\begin{center}
\includegraphics[width=0.9\linewidth]{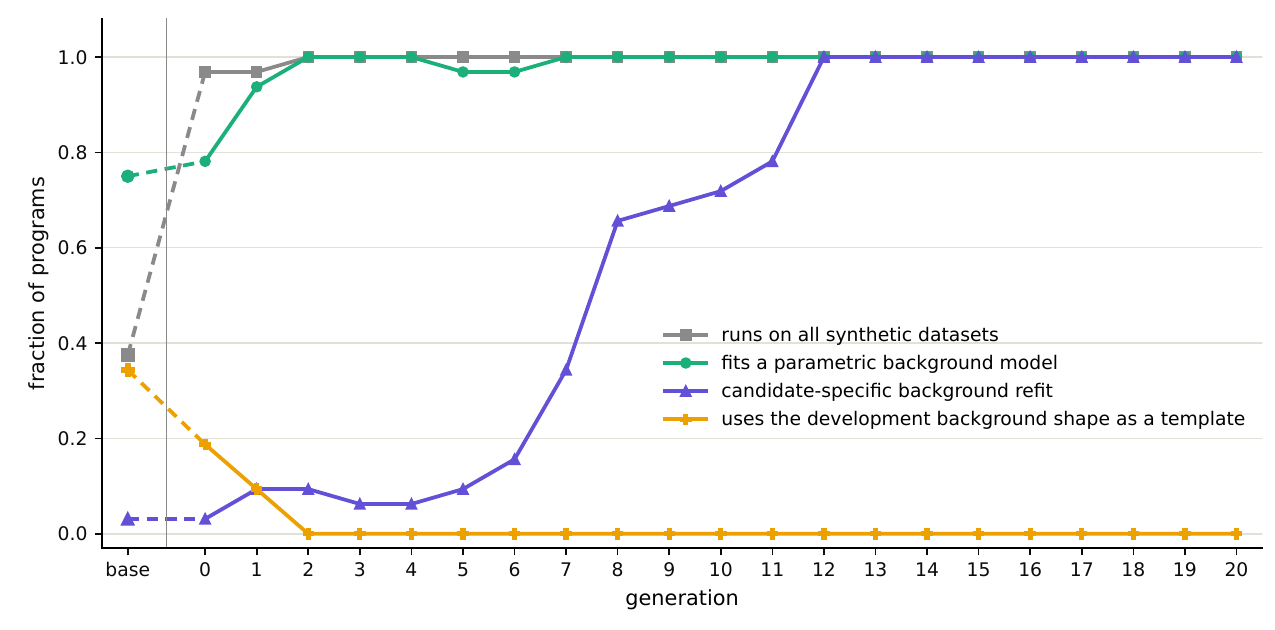}
\end{center}
\caption{Components of the 27B model's programs during evolution, as fractions of the 32 selected programs in each generation. Curves show execution success on all synthetic datasets (technical success rate), parametric background fitting (positive component), candidate-specific background refitting over the full mass window with a region around the candidate excluded (positive component), and use of the development background shape as a template (negative component).}
\label{fig:adoption}
\end{figure}

Evolution first improves technical reliability, then exposes trade-offs between physics objectives (Fig.~\ref{fig:adoption}). Execution success on synthetic data rises from 0.38 for base-model single shots to 0.97 at generation 0 and 1.00 from generation 2. On the twelve hidden datasets, early sensitivity gains increase false claims: the background-only score falls from 0.80 to 0.49 between generations 0 and 1, while the one- and two-resonance scores rise from 0.16 and 0.18 to 0.34 and 0.32. The background-only score then recovers to 0.90 at generation 12 as false claims become rare, while the one-resonance score continues to improve with more accurate peak localization. 
Two-resonance performance improves more slowly, leveling off at 0.41 for the selected population and 0.29 for the best-fitness program, against 0.47 for the reference. After generation 15, renewed losses on background-only data accompany rising synthetic fitness, indicating overfitting to the synthetic development dataset; the highest-fitness program nevertheless remains stable from generation 14.

Evolution produces a solution beyond the best initial random trial, with a reward of 0.644 against 0.578 for the best of 32 initial samples. 
The gain therefore reflects improvement beyond initial selection, with variation between runs.

The discovered method, documented in Appendix~\ref{app:program}, illustrates how good components are encouraged during evolution. The final 27B program fits an exponential-polynomial background, refits it with each candidate region excluded, and tests Gaussian peaks with free widths using a likelihood-ratio significance. Consistent with selection for higher reward (Fig.~\ref{fig:adoption}), parametric background fits rise from 0.78 of the population at generation 0 to 1.00 from generation 7, and candidate-specific refits from 0.03 to 1.00 by generation 12. Conversely, use of the development background as a template falls from 0.34 of base-model single shots to 0.19 at generation 0 and disappears from generation 2. Thus, reward-based selection encourages the model to estimate the background directly from the input dataset (analogous to real data), rather than relying on the background shape learned during development (analogous to simulation), without requiring external guidance or human-in-loop.

The gap between model sizes suggests that the ultimate limitation comes from the model capacity (Fig.~\ref{fig:steps27b}, Table~\ref{tab:results}). Although the 4B model improves its reward on synthetic datasets from 0.33 to 0.55, its selected programs fail to generalize to the twelve hidden datasets, consistently achieving only the trivial baseline reward of 0.333 from generation 6 onward. These programs rely on threshold-based searches over smoothed histograms without explicitly modeling the background, making them sensitive to the peak widths used in synthetic training. In contrast, the 27B model independently discovers a more robust strategy based on background estimation, highlighting a qualitative difference in the models' ability to develop generalizable solutions rather than merely optimize the training reward.
At sufficient capacity, however, reward-guided evolution already enables a useful search with the frozen base model.

\subsection{Reinforcement learning}
\label{sec:rl}

Evolution improves the search at a substantial cost: the 27B self-sampling and evolution stages executed about 2300 programs, 1300 of them in the 20 generations, and used about 2.9 node-hours on 4-GPU nodes plus an executor node. A new task would require the search to start again. 
We therefore update the model weights with the aim of reducing the iterations needed on similar and even new tasks. 

Supervised fine-tuning (SFT) first aligns the model to its best evolved programs through maximum-likelihood training, with each training pair containing the task prompt and one of the model's own programs selected by synthetic fitness. Across 32 single shots on the twelve hidden datasets, the 27B mean rises from 0.116 to 0.342, while the number scoring above 0.5 increases from 1 in 32 to 10 in 32 (Table~\ref{tab:results}), showing that the policy generates useful programs more frequently.
SFT nevertheless remains below the programs it imitates.
Maximum-likelihood training rewards reproduction of target code without directly evaluating the physics quality of generated programs; it therefore spreads imitation across the target family rather than concentrating on its best members. Execution success is 0.78 after SFT, and successful programs average 0.42, indicating that both technical failures and physics performance limit the policy.

We further apply Group Relative Policy Optimization (GRPO) to recover performance by optimizing the task reward directly. For each prompt, we execute a group of 8 sampled programs on synthetic datasets, reinforcing those above the group mean and suppressing those below it. By GRPO step 30, the 27B single-shot mean on the twelve hidden datasets recovers to 0.457 [0.40, 0.50], improving on SFT by $+0.12$ [$+0.09$, $+0.14$] and on the base model by $+0.34$ [$+0.28$, $+0.39$]. 
Most recovery comes from the background-only group, whose score rises from 0.64 to 0.94, while the one-resonance score increases from 0.20 to 0.26 and the two-resonance score does not improve (0.19 to 0.18).
The gain includes better physics performance as well as fewer failures.

The combined procedure therefore uses evolution to discover useful components and reward-based training to make them more frequent. 
As demonstrated by the 4B model, improvements in execution performance alone are insufficient when the generated programs lack a sound physics methodology, highlighting the limitations imposed by model capacity on intrinsic physics reasoning.
We reduce the number of trainable parameters and memory requirements by employing low-rank adaptation (LoRA)~\cite{hu2022lora}, while keeping the pretrained model weights frozen throughout training.

\begin{table}[t]
\centering
\footnotesize
\setlength{\tabcolsep}{5pt}
\caption{Rewards on the twelve hidden datasets, overall and by
resonance count (none, one, two). For $n=32$, values are the mean over
32 programs, with non-executable outputs counted as zero.
Values in parentheses give the best program of the cohort:
$^\dagger$ selected by synthetic fitness,
$^\ast$ selected by hidden reward (oracle diagnostic only).
Hidden rewards never enter training or development-time selection.}
\label{tab:results}

\begin{tabular}{@{}lrcccc@{}}
\toprule
Stage & $n$ & Reward & \multicolumn{3}{c}{Reward by resonance count} \\
\cmidrule(l){4-6}
 & & & none & one & two \\
\midrule

\multicolumn{6}{@{}l}{\emph{Qwen3.8-27B}} \\

30-min session, execution feedback
& 1 & 0.516
& 1.000 & 0.122 & 0.426 \\

Base, single shot
& 32 & 0.116 (0.578$^\ast$)
& 0.281 (1.000) & 0.029 (0.329) & 0.037 (0.404) \\

Evolution, generation 0
& 32 & 0.382
& 0.805 & 0.163 & 0.177 \\

Evolution, generation 12
& 32 & 0.626
& 0.898 & 0.584 & 0.397 \\

Evolution, generation 20
& 32 & 0.509 (0.644$^\dagger$)
& 0.477 (1.000) & 0.640 (0.642) & 0.411 (0.289) \\

SFT, single shot
& 32 & 0.342
& 0.641 & 0.195 & 0.189 \\

GRPO step 30, single shot
& 32 & 0.457 (0.683$^\ast$)
& 0.938 (1.000) & 0.255 (0.622) & 0.180 (0.427) \\

\midrule
\multicolumn{6}{@{}l}{\emph{Qwen3-4B-Instruct-2507}} \\

Base, single shot
& 32 & 0.013
& 0.031 & 0.000 & 0.008 \\

Evolution, generation 20
& 32 & 0.333 (0.333$^\dagger$)
& 1.000 (1.000) & 0.000 (0.000) & 0.000 (0.000) \\

GRPO step 50, single shot
& 32 & 0.333
& 1.000 & 0.000 & 0.000 \\

\midrule
\multicolumn{6}{@{}l}{\emph{References}} \\

Hand-written reference program
& 1 & 0.689
& 1.000 & 0.599 & 0.468 \\

Always abstain
& 1 & 0.333
& 1.000 & 0.000 & 0.000 \\

Always claim 125\,GeV
& 1 & 0.250
& 0.000 & 0.250 & 0.500 \\

\bottomrule
\end{tabular}
\end{table}

Table~\ref{tab:results} reports performance on held-out data following selection on synthetic data. All pool, evolutionary and training selections use synthetic data, while reported rewards use the twelve hidden datasets. 
The 27B policy substantially improves single-shot average performance, from 0.116 to 0.457, compared with 0.516 for its base model's 30-minute agentic session. Allowing oracle selection of the best among 32 single-shot outputs based on hidden test rewards would increase the reward to 0.683, approaching frontier-model out-of-box performance. However, this represents an optimistic upper bound rather than a valid evaluation protocol, as it assumes access to hidden test rewards (Fig.~\ref{fig:casestudy_benchmark}).
The 4B policy retains the limitation of its evolved programs and remains at the trivial baseline, consistent with the capacity gap observed during evolution.

\section{Conclusions}

We introduce a suite of open-ended collider physics challenges with continuous, automatically verifiable rewards, designed to both evaluate frontier AI models and provide environments for their improvement. Across six representative HEP tasks, we demonstrate the capabilities of current frontier models while identifying substantial room for improvement on challenging scientific problems. As an initial proof of concept, we demonstrate that reward-driven evolution and training can improve an open-weight model and encourage the development of more robust physics analysis strategies without human-provided solutions.

Looking ahead, we aim to expand this framework to a broader range of open scientific problems and more capable models, establishing increasingly challenging environments for autonomous learning and evaluation. We envision a shift from AI agents that automate existing scientific workflows toward agents that can independently develop, test, and improve scientific methodologies. By combining the rich challenges of fundamental physics with verifiable rewards, this approach offers a promising path toward self-improving AI for scientific discovery.

\section*{Acknowledgments}

This work is supported by the U.S. Department of Energy under contract number DE-AC02-76SF00515.

AI tools were used to assist with manuscript drafting and refinement, code development, data analysis, and figure generation. All results were reviewed and validated by the authors, who take full responsibility for the content and conclusions of this work.

\bibliography{colm2026_conference}
\bibliographystyle{colm2026_conference}

\appendix

\begin{table}[t]
\centering\footnotesize
\caption{Settings of the self-evolution and RL stages (27B / 4B where they differ).}
\label{tab:settings}
\begin{tabular}{@{}lp{4.2in}@{}}
\toprule
Stage & Setting \\
\midrule
Models & Qwen3-4B-Instruct-2507 and Qwen3.8-27B, base weights, no thinking. \\
\addlinespace[2pt]
Out of the box & 8 prompt variants $\times$ 4 samples; turn 1 = initial program, turn 2 = its revision after execution feedback; the paper reports turn 1. \\
\addlinespace[2pt]
Self-sampling pool & 256 chains $\times$ up to 4 turns with execution feedback; synthetic signals 4--8$\times$ the template; kept if reward $\geq 0.5$: 109 programs (27B), 46 (4B). \\
\addlinespace[2pt]
Evolution & population 32, 64 children per generation, 20 generations; 60\,\% mutation / 40\,\% crossover; fitness $=$ reward on two synthetic variants at real signal strength; 24 elites $+$ 8 random survivors. \\
\addlinespace[2pt]
SFT & LoRA rank 32, lr $10^{-4}$, 2 epochs; targets $=$ the model's own evolved programs with fitness $\geq 0.9$ (27B) / $\geq 0.55$ (4B): 149 programs $\times$ 2 prompts (27B), 76 $\times$ 2 (4B). \\
\addlinespace[2pt]
GRPO & from the SFT adapter, lr $10^{-5}$, 6 prompts $\times$ 8 rollouts per step, 30 steps (27B) / 50 steps (4B); reward $=$ score on the synthetic datasets of the prompt variant. \\
\addlinespace[2pt]
Evaluation & 32 single shots per model or policy on the twelve hidden datasets. \\
\bottomrule
\end{tabular}
\end{table}

\lstdefinestyle{prompt}{basicstyle=\ttfamily\scriptsize,breaklines=true,breakatwhitespace=false,
  columns=fullflexible,keepspaces=true,showstringspaces=false,frame=single,framerule=0.3pt,
  xleftmargin=0pt,breakindent=0pt,postbreak=\mbox{$\hookrightarrow$\space},escapechar={},escapeinside={}}

\section{Prompts used in the self-evolution pipeline}
\label{app:prompts}

All text shown in a framed block in this appendix is a verbatim copy of the repository source, extracted
by script from \path{HEP_RL/rl/peak_env.py}, \path{sample_multiturn.py}, \path{evolve.py} and
\path{hint_check.py}; long lines are wrapped at the arrow marks.
The same prompts are used for both policy models (Qwen3-4B-Instruct-2507 and Qwen3.8-27B); the pipeline
never names a method, and the base weights are frozen throughout the search.

\subsection{System prompt and hint-free task prompt}
\label{app:prompts:task}

Every episode starts from a two-message conversation: the system prompt \texttt{SYSTEM\_EN} and one user
message rendered from the template \texttt{TASK\_EN}; both are defined in \texttt{peak\_env.py},
lines 123--182. The model answers with one \texttt{python} code block, which is written to \texttt{search.py} and executed in a
\texttt{bwrap} sandbox on three freshly staged datasets (one with zero, one with one and one with two injected
resonances); the reward is the mean of the three per-dataset soft-F1 scores. The system prompt is:

\begin{lstlisting}[style=prompt]
You write Python programs for data-analysis tasks. Answer with exactly one ```python code block containing a complete, self-contained program. Keep any explanation outside the block to a few sentences.
\end{lstlisting}

The task prompt is the following Python format string (doubled braces \texttt{\{\{}~\texttt{\}\}} are
\texttt{str.format} escapes and render as single braces; the \texttt{\{...\}} fields are filled at run time as
explained below):

\begin{lstlisting}[style=prompt]
Task: build a reusable search for narrow diphoton resonances.

Each simulated proton-proton collision contains two photons and up to three jets.
A dataset is a full diphoton mass spectrum; its data.json declares a signal region
inside that spectrum. Decide how many narrow resonances the signal region holds
(zero, one or two) and report their masses. Any method is allowed, as long as it runs
on a CPU within the time limit and reads only its inputs.

Development data, in the directory dev-b/ next to your program (resolve the path
from your program's own location, not from the working directory):
- dev-b/data.root: a complete rehearsal dataset, the full spectrum (785,987 events);
  it holds one injected resonance (ggH at 125.0 GeV).
- dev-b/data.json: its metadata (same format as the datasets you will be run on).
- dev-b/signal_template.root: a weighted simulated sample of a narrow resonance
  (branch `weight`; the weights sum to the expected signal yield).
- dev-b/bkg_ideal.root: a weighted simulated background sample over the whole mass
  window (branch `weight`; the weights sum to the expected background yield).
The datasets you will be run on are not these files: their background shape and
yield, signal strengths and signal region differ. Read every boundary, luminosity
and selection from the dataset's own data.json at run time and hard-code nothing
from the development data.

Event format: a ROOT file with a TTree named events and 18 numeric branches:
pho1_pt pho1_eta pho1_phi pho2_pt pho2_eta pho2_phi
jet1_pt jet1_eta jet1_phi jet1_m jet2_pt jet2_eta jet2_phi jet2_m jet3_pt jet3_eta jet3_phi jet3_m
pt and mass in GeV, eta dimensionless, phi in radians. Photons are massless:
m_yy = sqrt(2 * pt1 * pt2 * (cosh(eta1 - eta2) - cos(phi1 - phi2))).
Missing jets are exact zero quadruplets. The datasets are unweighted event lists;
dev-b/signal_template.root and dev-b/bkg_ideal.root carry a `weight` branch.

Example data.json of a dataset (the signal region differs between datasets):
{
 "data": "data.root",
 "luminosity_fb": 139.0,
 "myy_window_gev": [
  105.0,
  145.0
 ],
 "signal_region_gev": [
  120.0,
  130.0
 ],
 "basis": "pt_eta_phi",
 "variables": [
  "pho1_pt",
  "pho1_eta",
  "pho1_phi",
  "pho2_pt",
  "pho2_eta",
  "pho2_phi",
  "jet1_pt",
  "jet1_eta",
  "jet1_phi",
  "jet1_m",
  "jet2_pt",
  "jet2_eta",
  "jet2_phi",
  "jet2_m",
  "jet3_pt",
  "jet3_eta",
  "jet3_phi",
  "jet3_m"
 ],
 "preselection": {
  "n_photons": 2,
  "photon_pt_over_myy_min": [
   0.35,
   0.25
  ],
  "photon_abs_eta_max": 2.37,
  "photon_abs_eta_excluded": [
   1.37,
   1.52
  ],
  "jet_pt_min_gev": 22.13,
  "jet_abs_eta_max": 4.4,
  "jet_slots": 3,
  "jet_padding": "an absent jet is the exact zero 4-vector, filled from slot 1 upward"
 }
}

Deliverable: one Python program, search.py, taking two positional arguments:
    python search.py DATASET_DIRECTORY OUTPUT_JSON
DATASET_DIRECTORY contains data.root and data.json for one full-spectrum dataset.
Write OUTPUT_JSON as {"peaks": [{"mass_GeV": 126.4}]} with zero, one or two finite
masses inside that dataset's half-open signal region; {"peaks": []} abstains.

Scoring: for each dataset the claims are matched one-to-one to the true resonances;
credit decays with the mass error on a 0.5 GeV scale and vanishes beyond 1 GeV, and
the dataset score is a soft F1 over true and claimed peaks. A background-only dataset
scores 1 for an empty claim and 0 for any claim. The program is run independently on
3 datasets holding zero, one or two resonances; a crash, timeout or invalid
output scores 0 for that dataset.

Constraints: CPU only, 2 threads, 60 seconds of wall-clock time per call
including reading the data; each call runs in a fresh working directory with no state
carried between calls; no network; available libraries: numpy, pandas, scipy,
scikit-learn and uproot 5 (read branches with
`uproot.open(path)["events"].arrays([...], library="np")`; there is no PyROOT `ROOT`
module and no uproot3/uproot4). Prompt variant: 0.
\end{lstlisting}

\section{The program found by the 27B self-evolution}
\label{app:program}

The program analysed here is the one with the highest fitness of the whole 27B run, a crossover child of generation 14, fitness 0.996 on the two fitness variants), and it
is selected on fitness alone. Its reward is 0.644 on the twelve hidden datasets (null/single/double
1.000/0.642/0.289). Table~\ref{tab:program_flow} lists its
eleven steps in execution order; the glossary below explains the techniques it combines. All parameter values
are read from the program source; the program never reads the dataset's truth and takes the signal-region
boundaries from the dataset's own \texttt{data.json}.

\providecommand{\cell}[1]{\begin{minipage}[t]{\linewidth}\raggedright #1\strut\end{minipage}}
\begin{footnotesize}
\begin{longtable}{@{}p{0.15\linewidth}p{0.38\linewidth}p{0.38\linewidth}@{}}
\caption{The eleven steps of the best-fitness 27B program. SR = signal region of the dataset; $\sigma$ =
Gaussian width; $m_0$ = candidate mass.}\label{tab:program_flow}\\
\toprule
Step & What it does & Key numbers \\
\midrule
\endfirsthead
\toprule
Step & What it does & Key numbers \\
\midrule
\endhead
\midrule
\multicolumn{3}{r}{\textit{continued on next page}}\\
\endfoot
\bottomrule
\endlastfoot
\cell{1 Inputs and selection} &
\cell{Read photon $p_T$, $\eta$, $\phi$; compute $m_{\gamma\gamma}=\sqrt{2\,p_{T1}p_{T2}(\cosh\Delta\eta-\cos\Delta\phi)}$;
apply the pre-selection declared in \texttt{data.json}.} &
\cell{$|\eta|<2.37$, crack $1.37$--$1.52$ vetoed, $p_T/m_{\gamma\gamma}\ge 0.35/0.25$; window 105--145~GeV.} \\\addlinespace
\cell{2 Signal-width prior} &
\cell{Try \texttt{signal\_template.root} in the dataset directory; fall back to a fixed width if absent. The
template is never in the dataset directory, so the fallback is always used. The peak widths of the hidden
datasets are 1.35--1.50~GeV, so the prior is too narrow by about 40\,\%.} &
\cell{$\sigma=\max(0.5,\,0.02\cdot 40)=0.8$~GeV (always the fallback).} \\\addlinespace
\cell{3 Binning} &
\cell{Fine binning tied to the assumed width, far below the detector resolution.} &
\cell{bin $=\sigma/12\approx 0.067$~GeV, about 600 bins.} \\\addlinespace
\cell{4 Global background} &
\cell{Fit $\exp(a_0+a_1 m+a_2 m^2)$ to the sidebands by least squares on $\log(\mathrm{counts})$ (not a
likelihood fit). A standard smooth continuum, no spurious-signal study.} &
\cell{Polynomial degree 2; degrees 3 and 4 are tried only if the degree-2 fit fails. Sidebands: more than 2~GeV outside the signal region, i.e.\ $m<116.5$ or $m>134.5$~GeV on the hidden datasets (the margin is $\max(2.5\,\sigma_{\mathrm{prior}},\,0.8~\mathrm{GeV})=2$~GeV).} \\\addlinespace
\cell{5 Candidate seeding} &
\cell{Residual = data $-$ background, boxcar-smoothed; robust noise estimate from the MAD; \texttt{find\_peaks}
on a descending threshold ladder, the first rung that returns candidates wins. If no rung does, the single
highest point is used when it exceeds $\max(2.5\,\sigma_{\mathrm{noise}},3)$.} &
\cell{18-bin (1.2~GeV) boxcar; $\sigma_{\mathrm{noise}}=1.4826\cdot\mathrm{MAD}$ over the whole window;
ladder $3\to2\to1.5\to1\times\sigma_{\mathrm{noise}}$; prominence $\ge 0.4\times$threshold; distance 9 bins
(0.6~GeV); search the signal region widened by 2~GeV on each side (116.5--134.5~GeV on the hidden datasets); keep the 10 highest.} \\\addlinespace
\cell{6 Local background} &
\cell{For each candidate, refit $\exp(\mathrm{poly})$ over the whole window with a region around the candidate
removed, so the bump does not inflate its own background; a second signal farther away stays inside the fit.} &
\cell{Exclude $|m-m_0|\le 2.4$~GeV ($3\times$ assumed $\sigma$); $m_0$ first clipped into the SR.} \\\addlinespace
\cell{7 Signal fit} &
\cell{Gaussian on the fixed local background, binned Poisson likelihood; initialised by a log-parabola, then a
width scan with the height solved analytically, then coordinate descent on $\mu$ and $\sigma$. The free
width partly compensates the 0.8~GeV prior.} &
\cell{Parabola to $\log(\mathrm{data}-\mathrm{bkg})$ over $\pm 4\sigma$ ($\pm 3.2$~GeV); 25 widths log-spaced
0.32--2.0~GeV; $\mu$ steps $\pm0.1\sigma$, $\pm0.3\sigma$; at most 40 rounds.} \\\addlinespace
\cell{8 Significance} &
\cell{Likelihood-ratio test statistic with the background held fixed (not profiled); local significance, no
look-elsewhere correction.} &
\cell{$Z=\sqrt{2\,(\mathrm{NLL}_{\mathrm{bkg}}-\mathrm{NLL}_{s+b})}$.} \\\addlinespace
\cell{9 Acceptance} &
\cell{(a) merge nearby candidates keeping the higher $Z$; (b) rank by AIC (the program calls it AICc); (c) accept on $Z$ and $\Delta$AIC;
(d) at most two peaks with a minimum separation. The $\Delta$AIC cut is algebraically $Z\ge3$ and never
binds; the separation rule ($\approx$1--2~GeV in practice) also blocks close pairs.} &
\cell{Merge within $\max(0.7~\mathrm{GeV},\,0.8\times\mbox{fitted }\sigma)$; accept if $Z\ge 3.5$ and
$\Delta\mathrm{AIC}\le-3$; separation $\ge\max(\mbox{fitted }\sigma\mbox{ of both},\,0.7~\mathrm{GeV})$.} \\\addlinespace
\cell{10 Mass estimate} &
\cell{Median of three estimators (fit $\mu$, positive-residual centroid, log-parabola vertex), a candidate outside
the SR being dropped; then averaged (median of two values) with an iterated Gaussian-weighted unbinned mean of
all events near the candidate, clipped to the SR. The unbinned mean includes background and any neighbouring
signal, so the falling continuum or a nearby peak can pull the claim by about 0.2~GeV.} &
\cell{Centroid within $\pm 2.5\times$ fitted $\sigma$; unbinned mean within $\pm 4w$,
$w=\max(0.8~\mathrm{GeV},\,\frac{1}{2}\,\mbox{fitted }\sigma)$, i.e.\ at least $\pm 3.2$~GeV.} \\\addlinespace
\cell{11 Output} &
\cell{Write the claims. The error handler tests \texttt{"sr\_lo" in globals()}, which is never true, so any
exception leads to an empty claim.} &
\cell{\texttt{\{"peaks": [\{"mass\_GeV": ...\}]\}}, 0--2 entries.} \\
\end{longtable}
\end{footnotesize}

\end{document}